\def\year{2022}\relax
\documentclass[letterpaper]{article} 
\usepackage{aaai22}  
\usepackage{times}  
\usepackage{helvet}  
\usepackage{courier}  
\usepackage[hyphens]{url}  
\usepackage{graphicx} 
\usepackage{natbib}  
\usepackage{caption} 
\DeclareCaptionStyle{ruled}{labelfont=normalfont,labelsep=colon,strut=off} 
\usepackage{algorithm}
\usepackage{algorithmic}
\usepackage{xcolor}

\usepackage{newfloat}
\usepackage{listings}
\usepackage{booktabs}
\floatstyle{ruled}
\newfloat{listing}{tb}{lst}{}
\floatname{listing}{Listing}

\title{On Semantic Cognition, Inductive Generalization, and Language Models}
\author{
    Kanishka Misra\footnote{The author gratefully acknowledges feedback from Julia Rayz and Allyson Ettinger.}
}
\affiliations{
    Purdue University\\

    West Lafayette, IN 47906\\
    kmisra@purdue.edu
}

\usepackage{bibentry}

\newcommand{\rnm}{R\&M}
\newcommand{\clutrr}{\textsc{clutrr}}

\begin{document}

\maketitle

\begin{abstract}
My doctoral research focuses on understanding semantic knowledge in neural network models trained solely to predict natural language (referred to as language models, or LMs), by drawing on insights from the study of concepts and categories grounded in cognitive science.
I propose a framework inspired by `inductive reasoning,' a phenomenon that sheds light on how humans utilize background knowledge to make inductive leaps and generalize from new pieces of information about concepts and their properties.
Drawing from experiments that study inductive reasoning, I propose to analyze semantic inductive generalization in LMs using phenomena observed in human-induction literature, investigate inductive behavior on tasks such as implicit reasoning and emergent feature recognition, and analyze and relate induction dynamics to the learned conceptual representation space.
\end{abstract}

\section{Introduction}
Humans often engage in `Inductive Reasoning,' the use of existing semantic knowledge to make inferences about novel cases.
For example, on encountering a new property - \textit{`Robins have T9 hormones,'} one might generalize (or project) it to all birds \citep{osherson1990category}. 
Inductive inferences are often domain dependent---e.g. biological information may be projected across a taxonomy (\textit{robin} and \textit{swan}), whereas behavioral information may project across a shared property (\textit{hawk} and \textit{tiger})---and may change during a human's development.
Inductive reasoning sheds light on the organization of human concept knowledge, and therefore plays an important role in theories of semantic cognition.

Meanwhile, computational advances in the field of natural language processing (NLP) have led to the development of complex neural network models of language (LMs).
LMs primarily represent language by encoding the distribution of words in their contexts from large corpora, using a process known as `pre-training.'
Their success on a range of higher level semantic tasks, combined with their black-box nature has given rise to a research program, the goals of which are to develop an understanding of the knowledge LMs gain from pre-training \citep{alishahi2019analyzing}.
In this thesis, I attempt to contribute to this goal by developing an inductive reasoning-based analysis framework to better understand the synthetic semantic cognition of LMs.

\section{Related Work}
As LMs grow larger, so does their capacity to retrieve memorized facts about the world (\textit{birds can fly}), leading to the development of the the paradigm known as `LMs as Knowledge Bases'.
This paradigm uses LMs to perform `commonsense reasoning' by simply querying pre-trained LMs with prompts that elicit word predictions corresponding to the retrieved fact \citep{petroni2019language}, or by fine-tuning and evaluating LMs on knowledge bases \citep{bosselut2019comet}.
This paradigm exclusively focuses on \textit{what} aspects of world knowledge are accessible through pre-training.
Such an inquiry sheds important light on the access to long-term semantic memory as it emerges from predicting words in context. 
I hope to extend this line of research by focusing on \textit{how} pre-trained LMs use semantic knowledge to process and generalize novel information, and to what extent their behavior aligns to that in humans.

Inspired by research in cognitive science, my thesis ties in the influential work of \citet[\rnm{}, henceforth]{rogers2004semantic}.
\rnm{} present a connectionist account of inductive reasoning, where they describe a feed-forward network that performs inductive projections of novel properties, and displays patterns comparable to inductions in children across multiple ages \citep{carey1985conceptual}.
Despite this connection, my thesis pursues a line of research independent to that of \rnm{} as it exclusively relies on representations learned by models from the statistics contained in language corpora.
It therefore targets the role played by language---more specifically, the pre-training of LMs to predict words in context---in facilitating the learning of semantic knowledge as opposed to the localist representations of \rnm{}, who make no such commitment.
More recently, \citet{sinha2019clutrr} introduced the \clutrr{} benchmark to study a different kind of inductive reasoning---one that is rooted in formal logic---in LMs on synthetic kinship information expressed as language.
Unlike \clutrr{}, the inductive reasoning capacities considered in my thesis make graded distinctions between generalizations across two different concepts -- i.e., generalization of a property to \textit{robin} may differ from that to \textit{penguins}.
Such distinctions are not considered in \clutrr{}, which instead focuses on discrete and logical generalization.

\section{Inductive Reasoning with Language Models}
I argue that a test of induction in LMs must satisfy two desiderata: (1) The LM must perform a semantically meaningful task that facilitates inductive reasoning -- i.e.,~during induction, the LM must accept novel information about concepts (\textit{robins can queem}), and then produce an output that can be used to conclude about how it applies the information to other concepts (\textit{canaries can queem} vs. \textit{giraffes can queem}); and (2) It must cast induction as a probabilistic inference -- induction experiments involve supplying humans with novel `premises', followed by an investigation of how likely they think a `conclusion' is, usually interpreted as a conditional probability.
Based on the aforementioned constraints, I propose to fine-tune the LMs under investigation to classify generic beliefs as true (\textit{cats have fur}) or false (\textit{dogs can fly}).
The beliefs will be sourced from a belief-bank, a data structure consisting of facts about the world, retrieved from existing commonsense knowledge-bases that represent concepts and their properties.
Novel information will be formed by linking existing concepts to properties consisting of nonce words (e.g., \textit{dax, fep, queem, etc.}).
During induction, the novel information will be processed by the LM using a standard backpropagation step, repeated until correct classification is achieved.
Next, the LM's weights will be frozen and the LM will be evaluated over a range of different conclusion statements by investigating its probability for the true label during a forward pass.

Using this general framework for simulating induction in LMs, I aim to answer the following research questions:

\paragraph{RQ1: What kinds of inductive generalizations about concepts are made by LMs?} 
To answer this question, I propose to devise induction experiments targeting findings from the human-induction literature \citep{osherson1990category, kemp2009structured}.
The goal of these experiments is to zero in on the kinds of semantic inductive biases represented in the LMs'~generalization capacities, and compare them to phenomena  that drive induction in humans.

\paragraph{RQ2: To what extent do LMs recognize and use emergent features during induction?} 
Capturing implicit knowledge is a long-standing goal for commonsense reasoning \citep{bosselut2019comet, talmor2020leap}.
This question targets whether LMs implicitly use features that emerge in the induction environment. 
For instance, the property of \textit{flight} is implicitly encoded in the set of concepts: \{\textit{robins, bats, airplanes}\}, which may share the same novel property during induction.
The objective here is to quantify the tendency of LMs to generalize the novel property to other concepts that possess the emergent feature (e.g., \textit{butterflies}).

\paragraph{RQ3: How do the inductive generalization capacities of LMs relate to their representational space?} 
In order to extend the interpretability of my methods and yield a mechanistic insight on how LMs characterize novel information, I propose to measure the correspondence between the induction dynamics (loss during induction backpropagation) of the LMs and their representational geometry.

Taken together, the above questions target a comprehensive exploration of induction in LMs, with experiments ranging from tests of hallmark phenomena in human-induction literature, to implicit reasoning and learning dynamics.

\section{Preliminary Work and Research Timeline}
In my previous research \citep{misra2021language}, I investigated induction in LMs using experiments that conform with the `LMs as knowledge bases' paradigm, with a focus on whether typicality of concepts (e.g., \textit{robins} are more typical \textit{birds} than are \textit{penguins}) manifests in the word prediction capacities of LMs.
My current research involves training LMs on the true/false task using data from existing knowledge-bases, as well as running preliminary experiments that target the presence of specific inductive biases in LMs (part of RQ1). Table \ref{tab:timeline} shows my research timeline.

\begin{table}[h]
\centering
\caption{Research Timeline}
\label{tab:timeline}
\begin{tabular}{@{}ll@{}}
\toprule
\textbf{Objective}        & \textbf{Timeline}                    \\ \midrule
RQ1              & October 2021 - February 2022 \\
Proposal Defense & January/February 2022       \\
RQ2              & March 2022 - August 2022   \\
RQ3              & September 2022 - January 2022 \\ 
Thesis Writing   & January 2023 - February 2023 \\\bottomrule
\end{tabular}
\end{table}
\bibliography{aaai22.bib}

\end{document}